\documentclass[sigconf]{aamas}  

\usepackage{booktabs}

\usepackage{flushend}
\setcopyright{ifaamas}  
\acmDOI{doi}  
\acmISBN{}  
\acmConference[AAMAS'20]{Proc.\@ of the 19th International Conference on Autonomous Agents and Multiagent Systems (AAMAS 2020), B.~An, N.~Yorke-Smith, A.~El~Fallah~Seghrouchni, G.~Sukthankar (eds.)}{May 2020}{Auckland, New Zealand}  
\acmYear{2020}  
\copyrightyear{2020}  
\acmPrice{}  

\renewcommand\footnotetextcopyrightpermission[1]{} 
\usepackage{todonotes}
\usepackage{bm}
\usepackage{subcaption}

\newcommand{\expect}{\mathbb{E}}

\newcommand{\aux}{\bm{\Psi}}

\newcommand{\loss}{\mathcal{L}}

\newcommand{\imgheight}{2.3in}

\begin{document}

\title{Work in Progress: Temporally Extended Auxiliary Tasks}  



%
\author{Craig Sherstan}
\authornote{This work was conducted while Sherstan was an intern at Borealis AI.}
\affiliation{%
 \institution{University of Alberta}
}
\email{sherstan@ualberta.ca}
\author{Bilal Kartal}
\affiliation{%
 \institution{Borealis AI}
}
\author{Pablo Hernandez-Leal}
\affiliation{%
 \institution{Borealis AI}
}
\author{Matthew E. Taylor}
\affiliation{%
 \institution{Borealis AI}
}
%
%
%
%
%
%

\begin{abstract}  
Predictive auxiliary tasks have been shown to improve performance in numerous reinforcement learning works, however, this effect is still not well understood. The primary purpose of the work presented here is to investigate the impact that an auxiliary task's prediction timescale has on the agent's policy performance. We consider auxiliary tasks which learn to make on-policy predictions using temporal difference learning. We test the impact of prediction timescale using a specific form of auxiliary task in which the input image is used as the prediction target, which we refer to as temporal difference autoencoders (TD-AE). We empirically evaluate the effect of TD-AE on the A2C algorithm in the VizDoom environment using different prediction timescales. While we do not observe a clear relationship between the prediction timescale on performance, we make the following observations: 1) using auxiliary tasks allows us to reduce the trajectory length of the A2C algorithm, 2) in some cases temporally extended TD-AE performs better than a straight autoencoder, 3) performance with auxiliary tasks is sensitive to the weight placed on the auxiliary loss, 4) despite this sensitivity, auxiliary tasks improved performance without extensive hyper-parameter tuning. Our overall conclusions are that TD-AE increases the robustness of the A2C algorithm to the trajectory length and while promising, further study is required to fully understand the relationship between auxiliary task prediction timescale and the agent's performance.
\end{abstract}


\maketitle


\section{Introduction}

In reinforcement learning (RL), an agent tries to learn how to act so as to maximize the amount of reward it will receive over the rest of its lifetime. RL agents are dependent on their representations of state --- their description, in vector form, used to describe the configuration of its external and internal environment.

\begin{figure}
    \centering
    \includegraphics[width=\linewidth]{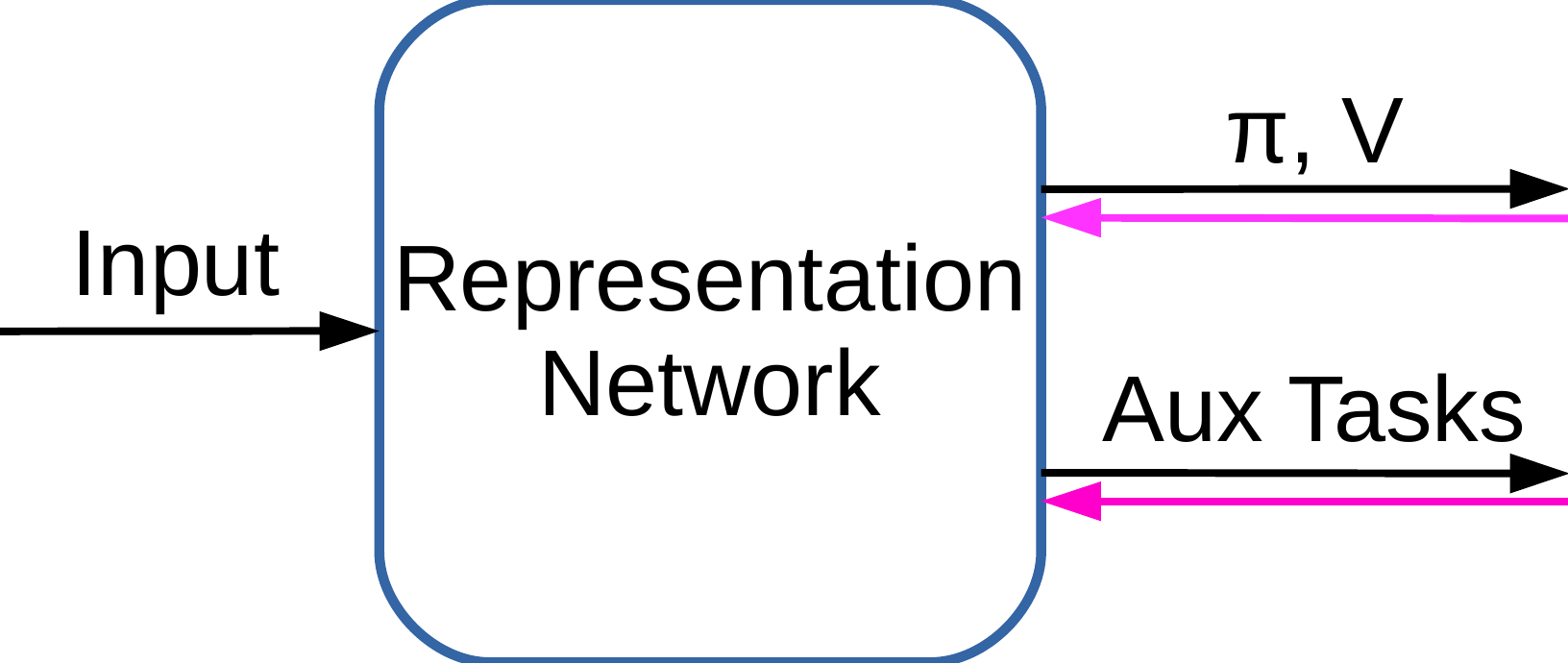}
    \caption{Auxiliary Tasks. Gradients (pink) flow back from both the primary tasks ($\pi$ and V) and the auxiliary tasks. In this paper, we assume that the output of the auxiliary tasks serve no purpose except to compute losses and provide gradients.}
    \label{fig:aux_tasks}
\end{figure}

In end-to-end training of deep reinforcement learning (DRL) the agent's representation, policy, and value estimates are trained simultaneously from reward received in the environment. Despite the successes in DRL, representation learning remains a serious challenge. A recent line of research is to use \emph{auxiliary tasks} to aid in driving the optimization process \citep{Jaderberg2017,Fedus2019,Le2018,Bellemare2016a,hern2019agent,kartal2019terminal,Veeriah2019}.
Let us define a task as any output of the network for which a loss function is attached. A \emph{primary task} is then any task that is directly related to an agent improving its policy. In value-based methods, this would include the value estimate, while in policy gradient methods this might include policy, value, and entropy losses (used to prevent early collapse of the policy). While what qualifies as an auxiliary task is open to interpretation, for the purposes of this paper let us define them as any task whose sole purpose is to assist in driving representation learning by providing additional gradients.
These ideas are illustrated in Figure~\ref{fig:aux_tasks} where both the primary tasks and auxiliary tasks depend on the same shared representation network and provide gradients for training the value-based reinforcement learner's network.

In this work we say that one representation is better than another if it enables either: 1) better final performance of the policy or 2) faster learning of the same policy. In this context, auxiliary tasks have been shown to both help and hinder performance \citep{Fedus2019}. At present, it is not well understood what tasks make good auxiliary tasks or why they help in general.

One example, where auxiliary tasks might help, is in the sparse reward setting, in which informative reward is infrequently observed. Auxiliary tasks may help by providing denser gradients. For example, an autoencoder auxiliary task \citep{Jaderberg2017} can provide training gradients on every training sample. Let us distinguish two cases then. In one case, the auxiliary task helps the network find the same representations as it would otherwise find, but faster. This may occur because the auxiliary task gradients drive the network towards the good representation faster or because they make the representation easier to learn (e.g., the additional gradients have the effect of reducing variance in the updates or make it easier to escape local minima \citep{suddarth1990rule}). In another case, these dense gradients impact the optimization process such that the representation found is in fact different, and possibly better. While we do not consider this setting here, a representation might also be considered ``better'' if it produces features that are more disentangled and generalize better in the transfer learning setting \cite{Le2018}. One of the results observed by \citet{Jaderberg2017} was that auxiliary tasks could make policy learning more robust to variations in learning parameters. We report on a similar effect in this paper.

\subsection{Hypothesis}

Numerous works have included predictive auxiliary tasks \citep{Jaderberg2017,Veeriah2019,Fedus2019,hern2019agent,kartal2019terminal}. However, the relationship between the prediction timescale and policy performance has not been studied. Our hypothesis is that having auxiliary tasks that predict the future on extended timescales will enable the agent to learn better policies. Further, we hypothesize that as the prediction timescale increases, from reconstruction towards the policy timescale of the agent, we should expect improved policy performance. This is motivated by the belief that on-policy temporally extended auxiliary tasks will share similar or even the same feature dependencies as the policy.

\subsection{Contributions}

To test our hypothesis we use temporally extended predictions of the agent's input image as auxiliary tasks, which we refer to as temporal difference autoencoders (TD-AE). We evaluate the impact of these auxiliary tasks on three scenarios in Vizdoom, a 3D game engine, using the A2C algorithm \citep{Mnih2016}. Our experiments do not conclusively support or disprove our hypothesis, however, we make several observations. Our first, and most notable observation, is that using TD-AE makes the learning more robust to the length of the trajectories, $n$, used in training. Reducing $n$ should allow the agent to adapt its policy more quickly and allows us to approach a more online algorithm. However, reducing $n$ from 128 to 8 steps causes the baseline algorithm to fail.
By including TD-AE auxiliary tasks the trajectory length can be reduced to 8 steps and still achieve performance which meets or exceeds that of the baseline. Our second observation is that, while we do not observe a direct relationship between auxiliary task timescale and policy performance, we do observe instances where the use of predictive auxiliary tasks outperform reconstructive tasks (i.e. TD-AE($\gamma=0$)). Our third observation is that the performance with auxiliary tasks is highly sensitive to the weighting placed on their losses. Finally, despite this sensitivity, the TD-AE auxiliary tasks are shown to generally improve performance even when the weighting is not well-tuned.

\section{Background}

In RL, an agent learns to maximize its lifetime reward by interacting with its environment. This is commonly modeled as a Markov Decision Process (MDP) defined by $\mathcal{S}, \mathcal{A}, P , r, \gamma$,  
where $\mathcal{S}$ is the set of states, $P$ describes the probability of transitioning from one state, $s$, to the next, $s'$, given an action, $a$, from the set of actions $\mathcal{A}$, that is $P(s'|s,a)$. The reward function generates a reward as a function of the transition: $r\equiv r(s,a,s')$. Finally, $\gamma\in[0,1]$ is a discount term applied to future rewards.

The agent's goal is to maximize the sum of undiscounted future rewards ($\gamma=1.0$), known as the \textit{return}. In practice we often use the discounted return instead:
\begin{align}
    G_{t} = R_{t+1} + \gamma R_{t+2} + \gamma^2 R_{t+3} + \ldots.
\end{align}

Measuring the goodness of being in a particular state and acting according to policy $\pi$ is given by the \textit{state-value}---the expected return from the state:
\begin{align}
    V_{\pi}(s) = \expect\bigg[G_t | S_t = s\bigg].
    \label{eq:value}
\end{align}
The goodness of taking an action from a state and then following policy $\pi$, known as the \textit{state-action value}, is given as:
\begin{align}
    Q_{\pi}(s,a) = \expect\bigg[G_t | S_t = s, A_t=a\bigg].
\end{align}
Temporal difference (TD) methods \citep{Sutton2018} update a value estimate towards samples of the discounted bootstrapped return by computing the TD error as:
\begin{align}
\delta_t = R_{t+1} + \gamma V(S_{t+1}) - V(S_t).
\end{align}
An $n$-step TD error, which is used in A2C, uses $n$ samples of the return before bootstrapping:
\begin{align}
    \delta_t = \bigg(\sum_{i=1}^{n}\gamma^{i-1}R_{t+i}\bigg) + \gamma^{n}V(S_{t+n}) - V(S_t).
    \label{eq:nstep}
\end{align}

\citet{Sutton2011} broadened the use of value estimation by introducing general value functions (GVFs). GVFs make two changes to the value function formulation. The first is to replace reward with any other measurable signal, which is now referred to as the \textit{cumulant}, $C$. The second change is to make discounting a function of the transition, $\gamma_t \equiv \gamma(S_t, A_t, S_{t+1})$ \citep{White2017}. Thus, the value function now estimates the expectation of the following return:
\begin{align*}
    G_t = C_{t+1} + \gamma_{t+1}C_{t+2} + \gamma_{t+1}\gamma_{t+2}C_{t+3}+\ldots.
\end{align*}

GVFs model elements of an agent's sensorimotor stream as temporally extended predictions. They can be learned using standard temporal difference learning methods \citep{Sutton2018}.


Advantage actor-critic (A2C) \citep{Mnih2016} is an algorithm that employs a parallelized synchronous training scheme (e.g., using multiple CPUs) for efficiency; it is an on-policy RL method that does not use an experience replay buffer. In A2C multiple agents simultaneously accumulate transitions using the same policy in separate copies of the environment. When all workers have gathered a fixed number of transitions the resulting batch of samples is used to train the policy and value estimates. A2C maintains a parameterized policy (actor) $\pi(a|s;\theta)$ and value function (critic) $V(s; \theta_v)$. The value estimate is updated using the $n$-step TD error (Eq.~\eqref{eq:nstep}).
It is common to use a softmax output layer for the policy head $\pi(A_t|S_t; \theta)$ and one linear output for the value function head $V (s_t; \theta_v)$, with all non-output layers shared (see Figure~\ref{fig:aux_tasks}).  
The loss function of A2C is composed of three terms: policy loss, $\loss_\pi$, value loss, $\loss_v$ and negative entropy loss, $\loss_H$. Entropy is treated as a bonus to discourage the policy from prematurely converging; the policy is rewarded for having high entropy. Each of the losses is weighted by a corresponding scalar, $\lambda$. Thus, the A2C loss is:

\begin{align*}
    \loss_{A2C} = \loss_{\pi} + \lambda_v\loss_v + \lambda_H\loss_H,
\end{align*}

where each loss component is weighted by a scalar $\lambda$ with $\lambda_v=0.5$, $\lambda_H=0.001$.

\section{Temporally Extended Auxiliary Tasks}

Consider writing value in the following form:

\begin{align*}
	V_\pi(s) = & \expect[R_{t+1}] + \gamma \expect[R_{t+2}] + \gamma^2\expect[R_{t+3}] + \ldots \\
	& + \gamma^{T-t-1} \expect[R_{T}] + \gamma^{T-t}V(S_T),
\end{align*}

\noindent
where each expectation is conditioned on $S_t=s$ and taken over the policy $\pi$. The value function does not directly observe state $s$, but instead uses a feature vector $\phi$ to describe the state, thus, $\phi_t\equiv\phi(S_t)$. Let us further say that to predict the expected reward $k$ steps in the future requires some feature vector $\phi(S_t;k)$, which contains all information required. That is, to accurately predict $\expect[R_{t+k}|S_t=s]$ requires all the information represented by $\phi(S_t;k)$. 
While $\phi$ might be independent for all $k$, $V(s)$ depends on the entire set of $\phi$ for all $k$. Thus, for a given cumulant and policy, value estimates at all timescales depend on the same features and the gradients produced during optimization should ideally lead to representations which capture all the same feature dependencies as required by the policy. 

However, the discounting term $\gamma$ places more emphasis on near term rewards (for all $\gamma < 1.0$), thus the features for near term rewards have a stronger impact on the value estimates performance. This dependency is further modulated by the magnitude of the rewards at each timestep in the future.
Thus, the optimization will focus on some of these features over the others. Therefore, given that our RL agents care about reward off into the future---typical values of $\gamma$ are $0.9, 0.95, 0.99$---it is reasonable to think that auxiliary tasks that share the same temporal feature dependencies would drive the network towards the same representation. 


\section{Methods}
\label{sec:methods}

To study the effects of the auxiliary task temporal prediction length on representation learning, we first need to define prediction targets. Here we simply choose the image input, $X_t$, as the prediction target, giving us the GVF defined as:

\begin{align}
    \aux_\pi(s) &= \expect_\pi\bigg[X_t + \gamma \aux(S_{t+1}) | S_t = s\bigg].
\end{align}

This requires a small change in the GVF definition. While the cumulant is usually defined as a function of the transition, \\
$C_{t+1}(S_t, A_t, S_{t+1})$, here, instead, the cumulant is only dependent on the start of the transition and is thus subscripted by $t$ (similar to the form used for defining the successor representation \citep{Sherstan2018c}). This GVF, which we refer to as TD-AE($\gamma$), has the form of a temporally extended autoencoder. Note that when $\gamma=0$ the target is simply the reconstruction of the input, i.e., an autoencoder. This GVF can be learned using standard TD methods and throughout the experiments in this paper we use TD(0), where the loss is:

\begin{align*}
    \delta_t^2 = (X_t + \gamma \aux(S_{t+1}) - \aux(S_t))^2.
\end{align*}

TD-AE is a conceptually unusual prediction target. When the input is an image, the sum of future discounted images has no obvious interpretation. However, we can instead think of each pixel in the image as a GVF cumulant, in which case the TD-AE is a collection of a large number of GVFs with well defined target signals. Each GVF predicts the sum of discounted future values for a certain pixel. In the GVF setting this is straightforward and does not require an easily interpreted description.

\citet{Sherstan2020} proposed a method for scaling TD losses by their timescale such that losses and corresponding gradients for different timescale value estimates are approximately of the same magnitude. We use this scaling here, which involves two modifications. The first is to rescale the network output by explicitly dividing by $(1-\gamma)$. The second is to multiply the TD error by the same factor. An average loss is taken across all $d$ pixels of the image. Thus, in our experiments, the loss used for the TD-AE is computed as follows:
\begin{align*}
\loss_{TD-AE} = \frac{1}{d}\sum_{i}^d[(1-\gamma)(X_{t:i} + \gamma \aux_i(S_{t+1}) - \aux_i(S_t))]^2.
\end{align*}

Combining this loss with the A2C loss gives:
\begin{align*}
    \loss = \loss_{A2C} + \lambda_{TD-AE}\loss_{TD-AE}
\end{align*}

$\lambda_{TD-AE}$ is a parameter we sweep over in our experiments.

\begin{figure}[t!]
    \centering
    \includegraphics[height=\imgheight]{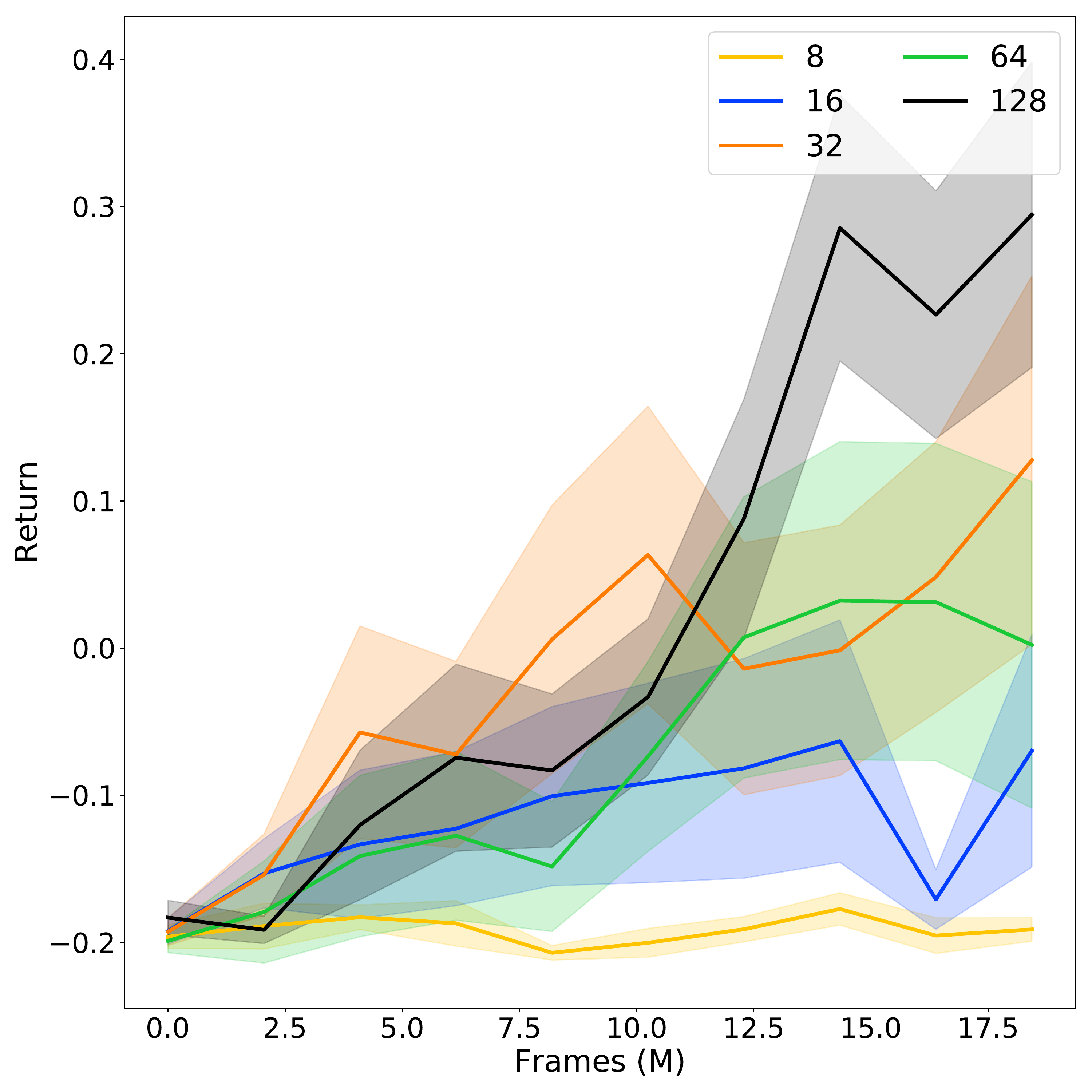}
    \caption{The effect of sequence length on performance  on the K-Item 2 scenario in Vizdoom. Legend indicates the length of the sequence length used for training. As the sequence length reduces the baseline becomes increasingly likely to fall into a failure mode.}
    \label{fig:effect_of_n}
\end{figure}


\begin{figure*}[t!]
\begin{center}
\begin{subfigure}[t]{0.32\linewidth}
    \centering
    \includegraphics[width=\linewidth]{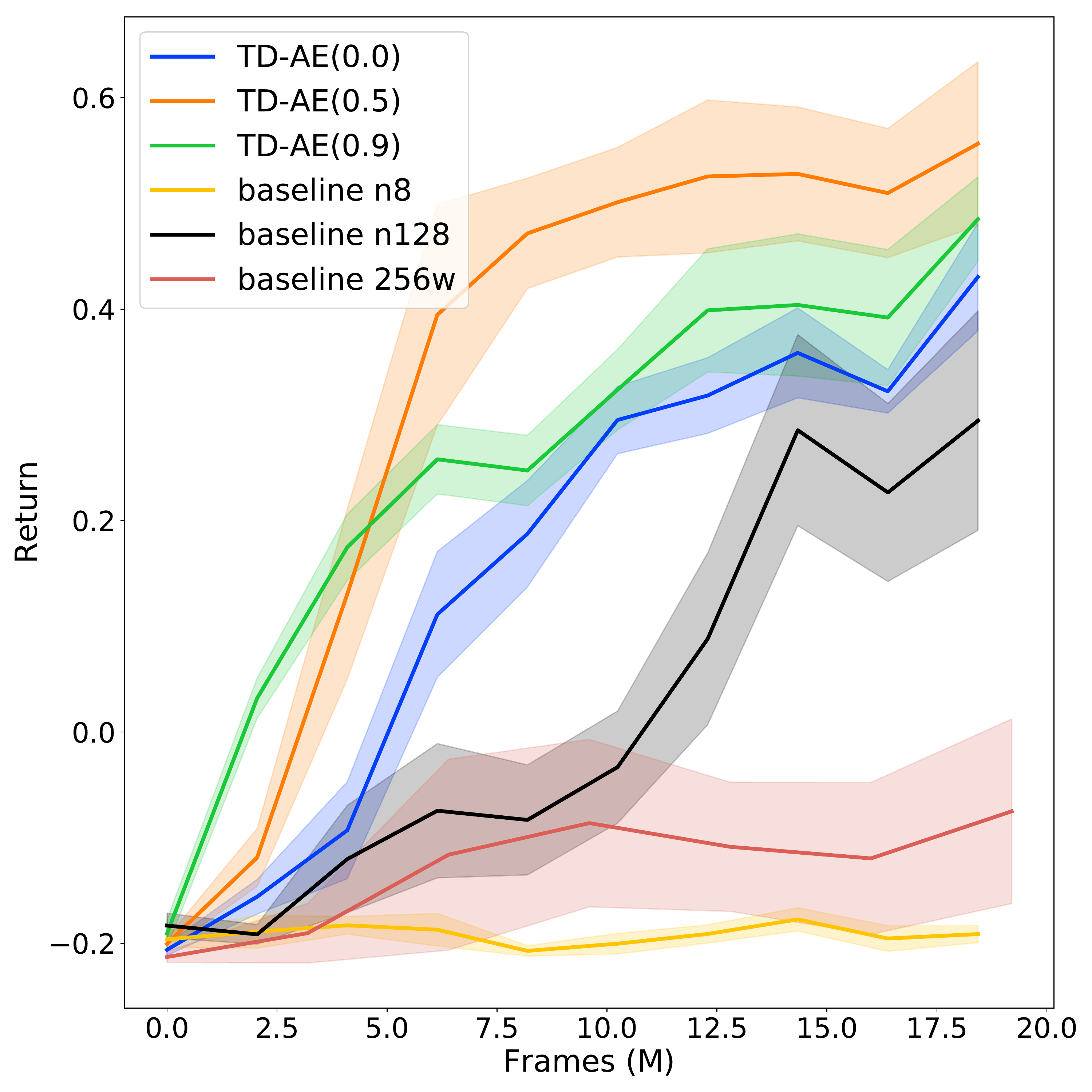}
    \caption{K-Item 2}
    \label{fig:kitem2}
\end{subfigure}
\hfill
\begin{subfigure}[t]{0.32\linewidth}
    \centering
    \includegraphics[width=\linewidth]{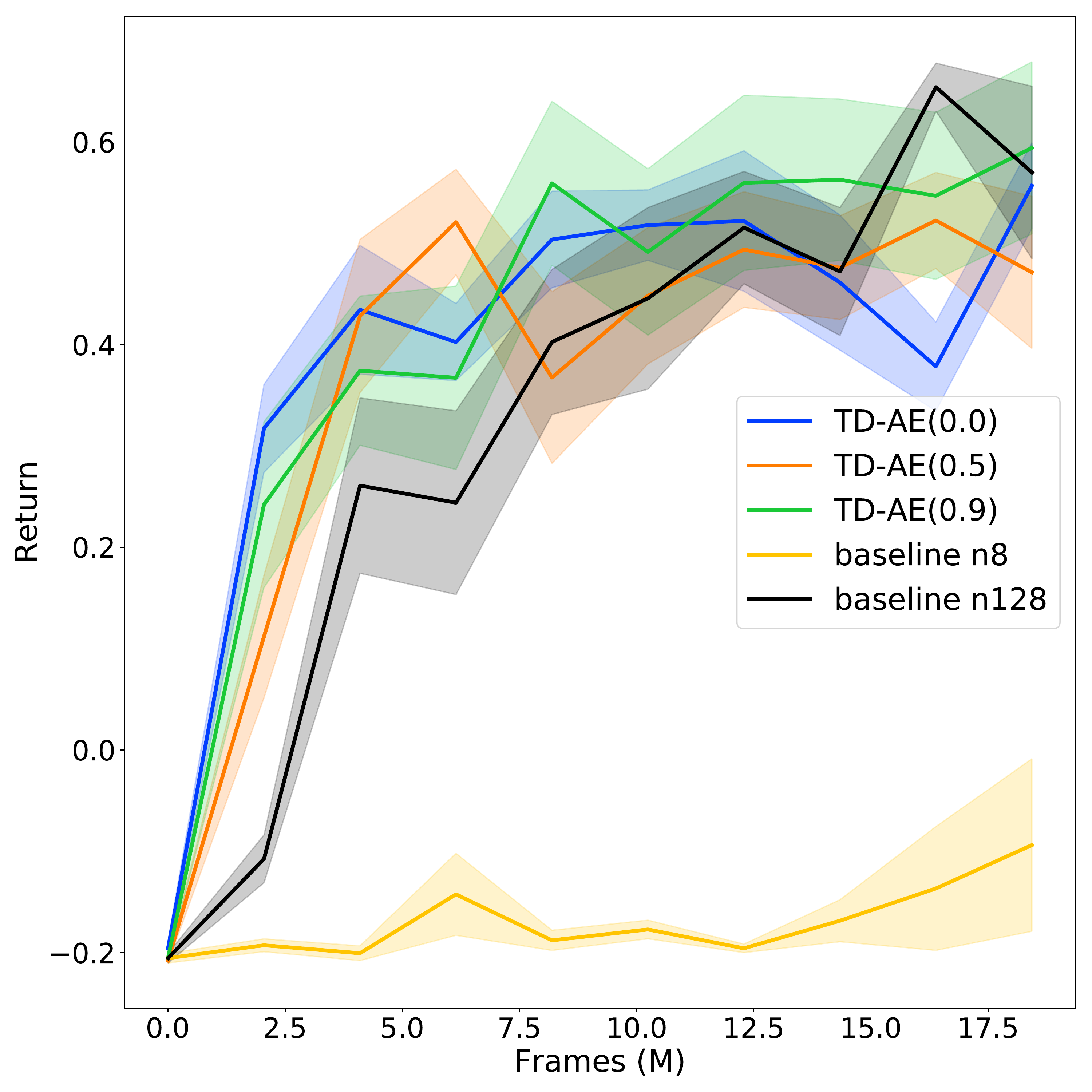}
    \caption{Labyrinth 13}
    \label{fig:labyrinth13}
\end{subfigure}
\hfill
\begin{subfigure}[t]{0.32\linewidth}
    \centering
    \includegraphics[width=\linewidth]{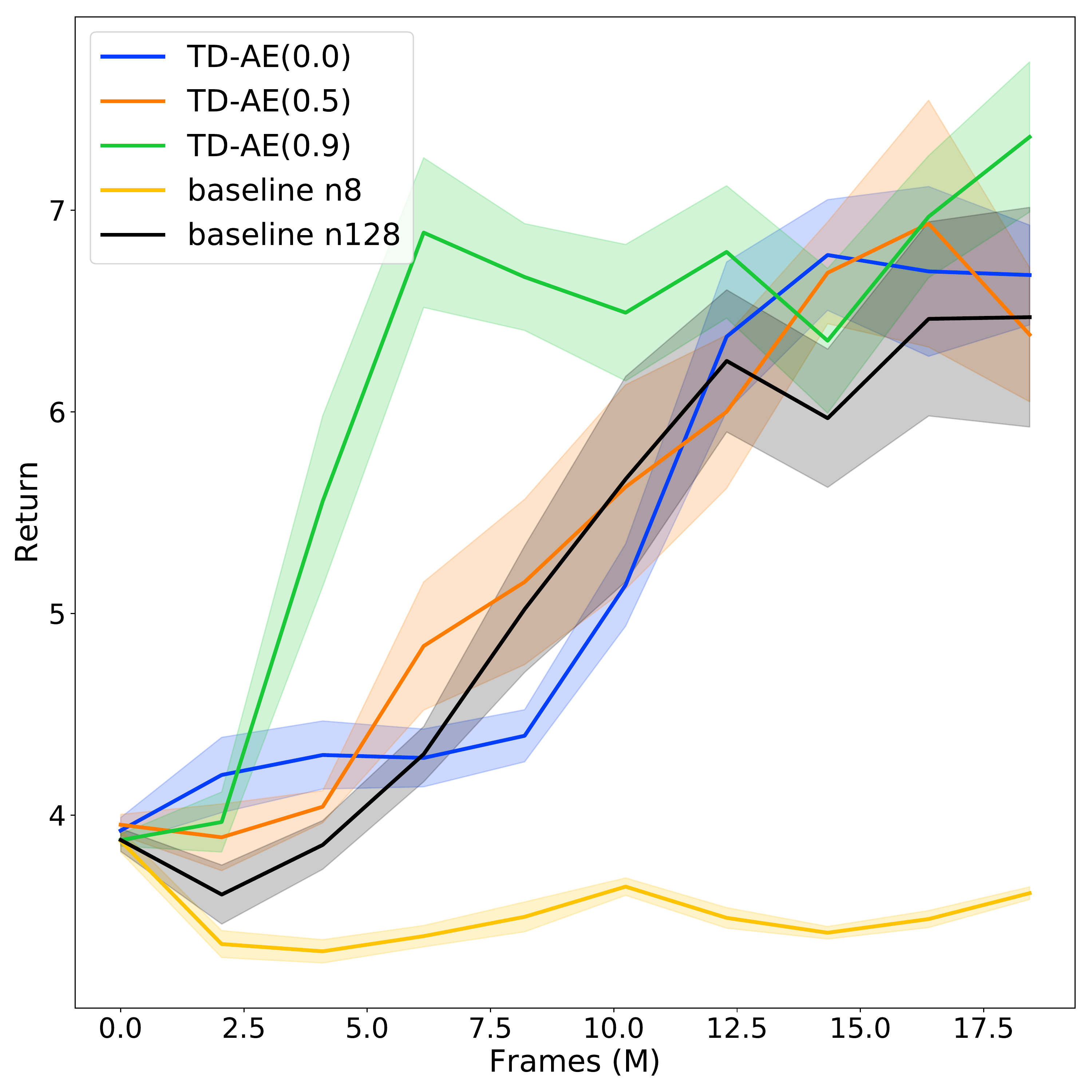}
    \caption{Two Color 5}
    \label{fig:twocolor5}
\end{subfigure}
\end{center}
\caption{Performance with and without TD-AE auxiliary task on three Vizdoom scenarios. The original baseline with n=128 is given in black. The baseline with n=8 is given in yellow and we see that it collapses and is not able to learn. Adding the TD-AE restores the performance of the network and in some cases even outperforms the original baseline. For the performance of the various TD-AE we chose the best $\lambda_{TD-AE}$ from our sweeps (Table~\ref{table:lambda_tdae}).}
\label{fig:aux_performance}
\end{figure*}

\section{Experiments}

\begin{figure*}[t!]
\begin{center}
\begin{subfigure}[t]{0.32\linewidth}
    \centering
    \includegraphics[width=\linewidth]{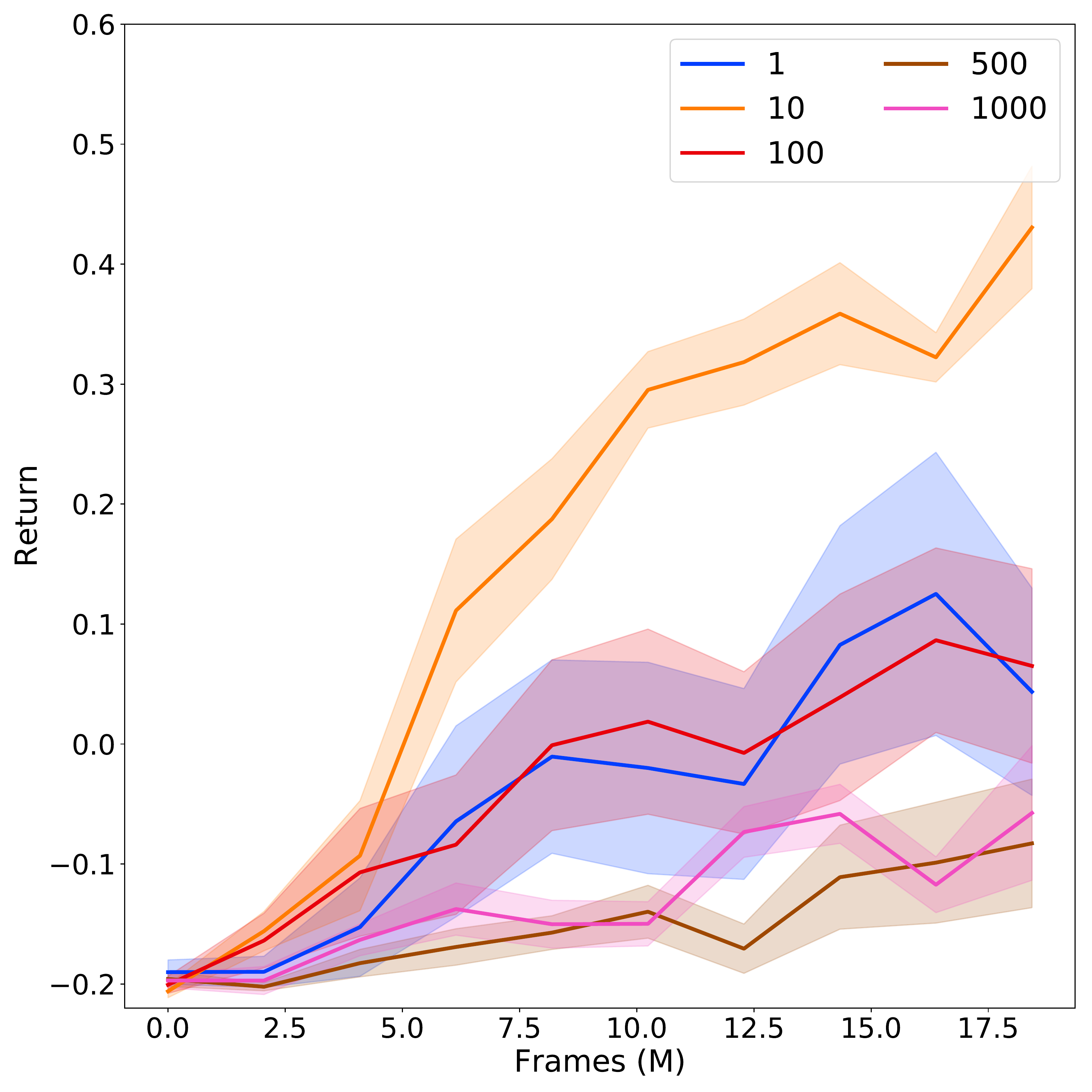}
    \caption{TD-AE(0.0)}
    \label{fig:kitem2.weighting.g_0_0}
\end{subfigure}
\hfill
\begin{subfigure}[t]{0.32\linewidth}
    \centering
    \includegraphics[width=\linewidth]{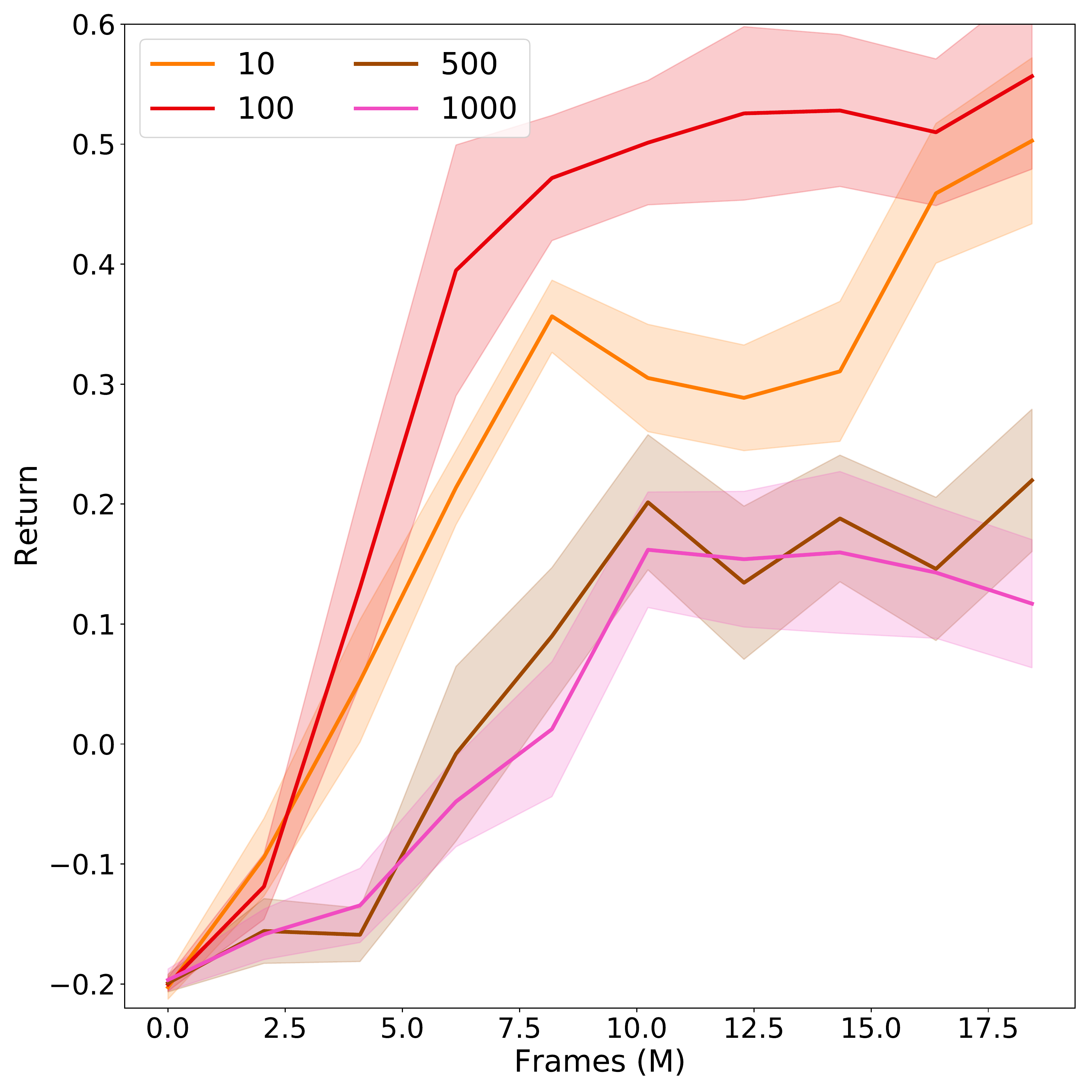}
    \caption{TD-AE(0.5)}
    \label{fig:kitem2.weighting.g_0_5}
\end{subfigure}
\hfill
\begin{subfigure}[t]{0.32\linewidth}
    \centering
    \includegraphics[width=\linewidth]{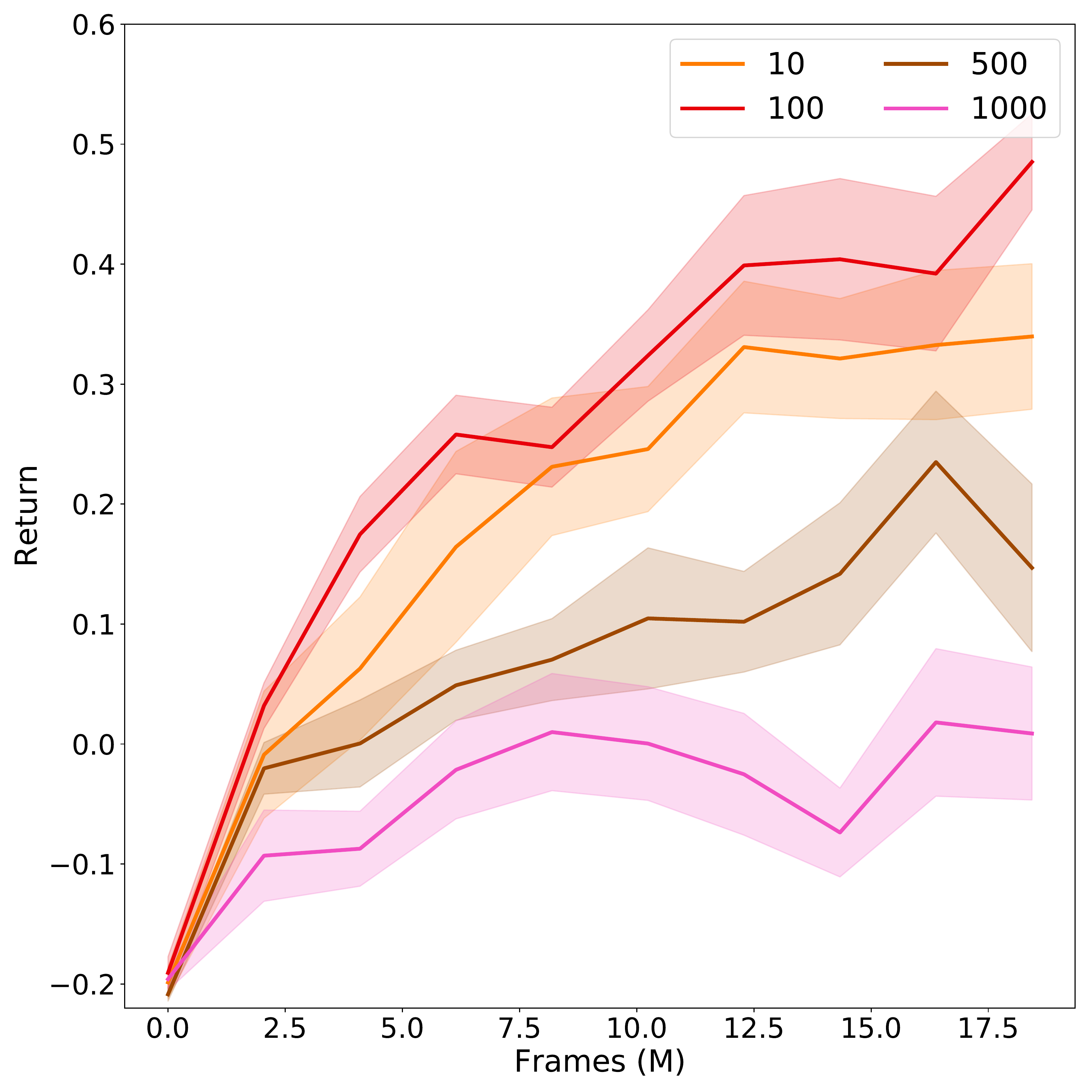}
    \caption{TD-AE(0.9)}
    \label{fig:kitem2.weighting.g_0_9}
\end{subfigure}
\end{center}
\caption{Effect of auxiliary loss weighting on the \textit{K-Item 2} task.}
\label{fig:kitem2.weighting}
\end{figure*}

\subsection{Setting}
Vizdoom \citep{Kempka2016} is a customizable 3D first-person environment. For evaluation in the Vizdoom environment, we used the A2C base code from \citet{Beeching2019}. In our experiments, 16 agents run simultaneously collecting transition samples that are used for batch training of the agent. The baseline uses an update sequence of 128 steps. Thus, each training batch consists of 2048 transition tuples obtained from 16 sequences of 128 steps. The agent's policy's discount was $\gamma=0.99$.

The core network consists of 3 convolutional layers followed by a fully connected layer with each layer using ReLU activation. This is fed into a GRU memory layer whose output is used as input to the policy, value, and TD-AE heads. The decoder for the TD-AE is composed of 3 fully connected layers with sigmoid activations. The sizes of the first two layers are 256 and 512 and the final layer, which has a linear activation, outputs the same size and shape as the original input images (64 $\times$ 112 pixels with 3 color channels). 

Each experiment was run over 10 seeds for 20 million frames (using 16 CPUs $\approx$ 10 hrs per run). Our figures use shading to indicate the standard error. While 10 seeds is not enough to give statistical certainty of the mean, we use standard error here for display purposes. Thus, shading should be considered as a scaled indication of the variability about the mean. Evaluations for the plots were performed by intermittently running test phases (approximately every 2 M frames). For each test phase 50 games were run with frozen network weights and the average episode return was reported.

Multiple Vizdoom scenarios were investigated, including \textit{K-Item 2, Labyrinth 13}, and \textit{Two Color Maze 5} (see descriptions in \citet{Beeching2019}). These scenarios were chosen as they required memory of the past and were thus expected to have long-term dependencies.

\subsection{Reducing Sequence Length \textit{n}}
We observe a consistent behavior in all of the tasks: when we reduce the length of the update sequence performance suffers (Figure~\ref{fig:effect_of_n}). 
However, adding an auxiliary task can restore baseline performance and in some cases exceed it. Figure~\ref{fig:aux_performance} shows this behavior across different Vizdoom scenarios. The default performance, with no auxiliary tasks and a sequence length of 128, is listed as \textit{baseline~n128}. In all tasks decreasing the sequence length to $n=8$, \textit{baseline~n8}, significantly reduces performance. Adding in the TD-AE consistently restores performance, and in some cases exceeds the performance observed on \textit{baseline~n128}. The performance reported for each series uses the best $\lambda_{TD-AE}$ found in our sweep. While we do not see a clear relationship between the timescale of the TD-AE prediction we do observe that in two of the scenarios using a $\gamma>0$ outperforms an autoencoder, $\gamma=0.0$ (Figures~\ref{fig:kitem2},\ref{fig:twocolor5}). 

Reducing the sequence length has the effect of reducing the batch size. In Figure~\ref{fig:kitem2} we show an additional experiment, \textit{baseline 256w}, which used no auxiliary tasks, $n=8$, and 256 workers, producing the same batch size as \textit{baseline n128}. Simply restoring the batch size did not restore the performance.


\begin{table}[t!]
    \caption{Values of $\lambda_{TD-AE}$ swept over. Optimal value for each $\gamma$ is indicated in bold. The size of these weights are much larger than reported in other works; this is due to the loss scaling of $1-\gamma$ we described in Section~\ref{sec:methods}}
    \label{table:lambda_tdae}
    \centering
    \begin{tabular}{|c|c|}
        \hline
        $\gamma$ & $\lambda_{TD-AE}$ \\
        \hline
        \multicolumn{2}{|c|}{\textbf{K-Item 2} and \textbf{Labyrinth 13}} \\
        \hline
        0.0 & \{1, \textbf{10}, 100, 500, 1000\} \\
        0.5, 0.9 & \{10, \textbf{100}, 500, 1000\} \\
        \hline
        \multicolumn{2}{|c|}{\textbf{Two Color 5}} \\
        \hline
        0.0, 0.5, 0.9 & \{10, 100, \textbf{500}, 1000\} \\
        \hline
    \end{tabular}

\end{table}


\subsection{Sensitivity to Loss Weighting}
We swept over $\lambda_{TD-AE}$ (see Table~\ref{table:lambda_tdae}) and found the performance of the policy could be sensitive to this term. This is shown in Figure~\ref{fig:kitem2.weighting} which depicts the performance on the \textit{K-Item 2} scenario for different $\gamma$ and weighting. Too little or too much weighting reduced performance. This is consistent with results found by others \citep{hern2019agent}. However, we note that in some works the same weighting is used for all scenarios. This could be problematic as our results showed that the same weighting was not optimal in all our scenarios.
We might speculate on the reason for this sensitivity as follows. If there is too little weight the resulting gradients are too small and result in smaller update steps, making it slower to learn the policy and decreasing the likelihood of escaping local minima or failure modes. On the other hand, if there is too much weight then the gradients of the auxiliary losses might interfere with those of the policy \citep{Schaul2019}. 

\subsection{Bi-Modal Performance}
\begin{figure}[t!]
    \centering
    \includegraphics[height=\imgheight]{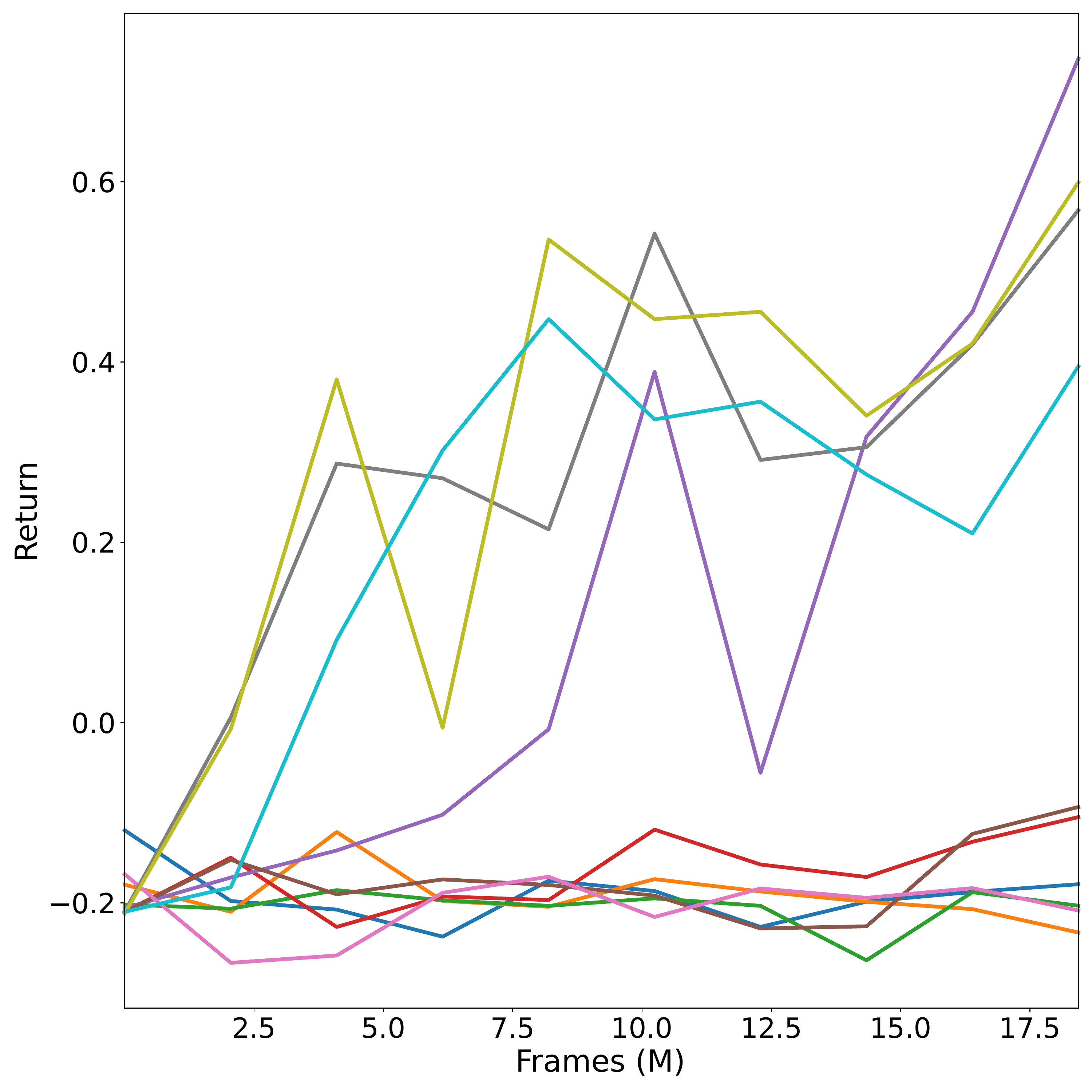}
    \caption{The mean and standard error can hide the fact that the returns split into two different modes (K-Item2, baseline n32). Each series indicates a different random seed. Here six of ten seeds are in the failure mode.}
    \label{fig:multimodal}
\end{figure}

In the plots we have shown the mean of multiple runs along with the standard error. However, this hides the underlying behavior. The performance of the agent would generally fall into two distinct modes. In the first mode, which we'll refer to as the learning mode, the performance curve continues to improve with more training, for example \textit{baseline n128}. In the second mode, which we refer to as the \textit{failure} mode, learning quickly plateaus and generally does not seem to recover, for example \textit{baseline n8}. 
Figure~\ref{fig:effect_of_n} suggests that as we decrease $n$, the likelihood of achieving the learning mode is reduced. Thus, simply presenting the mean masks the underlying distribution of returns like those shown in Figure~\ref{fig:multimodal}, which shows the performance of each seed for \textit{baseline n32}. We see that six out of the ten seeds resulted in the failure mode. Thus, the mean underrepresents those times when the policy learning succeeds and over represents the failure mode.

\subsection{TD-AE Predictions} 
\begin{figure*}[t!]
\begin{center}
\begin{subfigure}[t]{0.49\linewidth}
    \centering
    \includegraphics[width=\linewidth]{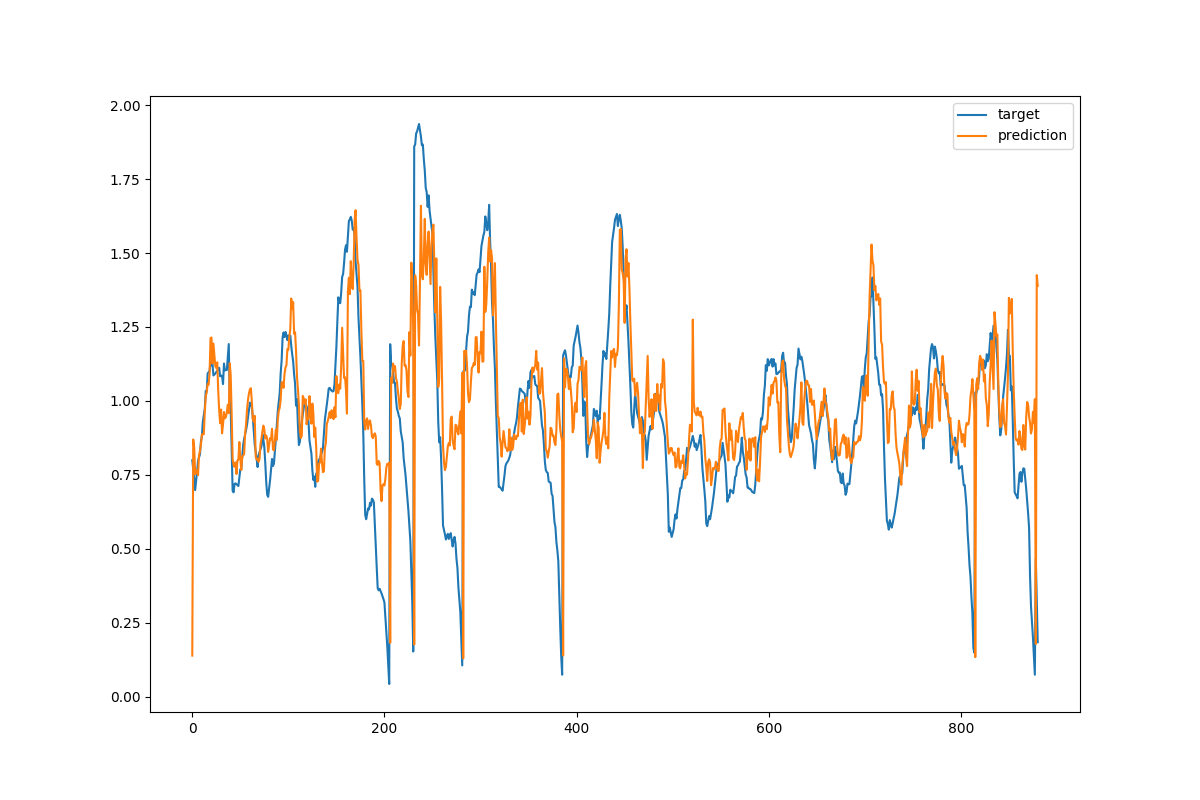}
    \caption{}
    \label{fig:aux_good}
\end{subfigure}
\hfill
\begin{subfigure}[t]{0.49\linewidth}
    \centering
    \includegraphics[width=\linewidth]{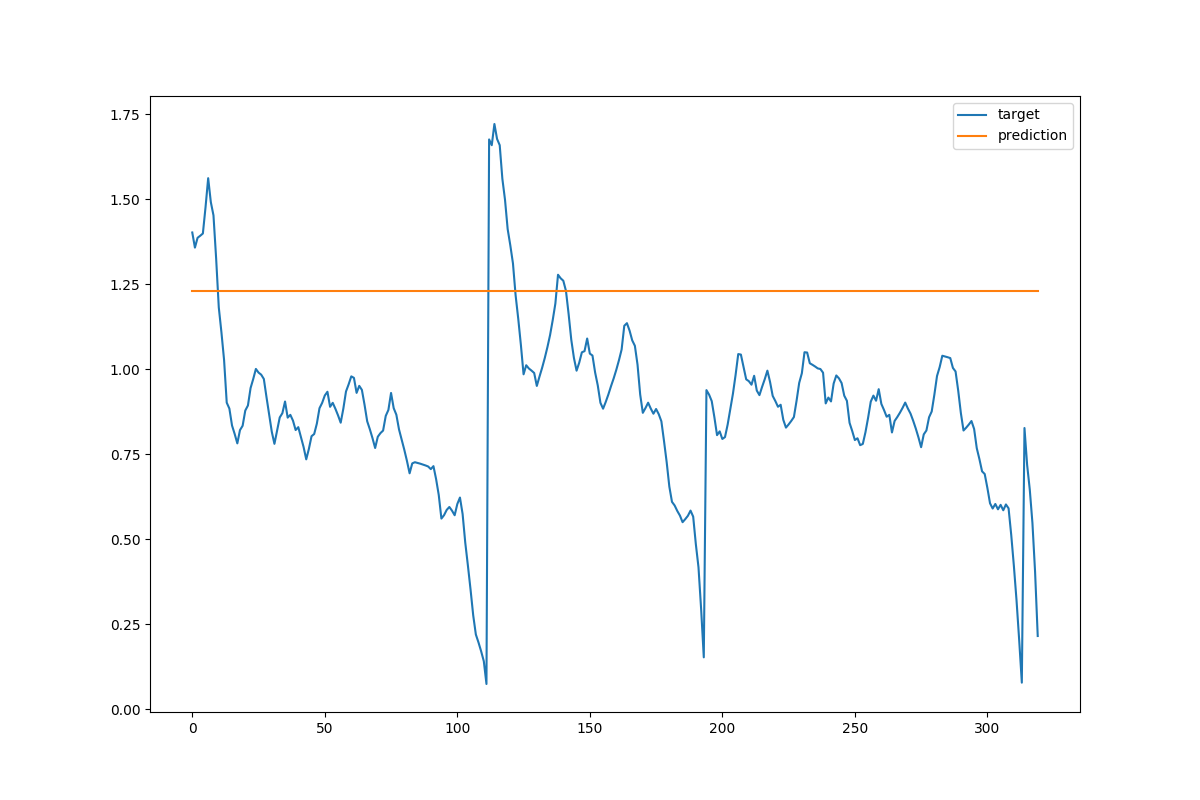}
    \caption{}
    \label{fig:aux_bad}
\end{subfigure}
\end{center}
\caption{Prediction quality for individual pixels for TD-AE($0.9$) (Two Color 3). For some pixels the prediction (orange) matches the target (blue) very well as in (a), while for other pixels the prediction does very poorly as in (b).}
\label{fig:aux_good_bad}
\end{figure*}

Figure~\ref{fig:aux_good_bad} considers auxiliary predictions of two different pixels over time. Transitions were gathered sequentially starting from an episode reset and potentially spanning several task terminations and resets. The true return is shown in blue and the predictions of TD-AE(0.9) are given in orange. While some predictions matched the target very well others failed (e.g., in Figure~\ref{fig:aux_bad}, the prediction is always a constant value).

\section{Discussion and Future Work}

We set out to test the hypothesis that temporally extended auxiliary tasks improve policy learning. We expected that as the timescale of the auxiliary task became longer, improvement would increase. If this hypothesis held true it would give some direction on the kinds of auxiliary tasks to use in a learning system. 

We observed that even autoencoders (TD-AE(0.0)), a commonly used auxiliary task \citep{Le2018, Shelhamer2016}, significantly improved learning.
Further, in some cases, we saw that TD-AE with $\gamma>0$ had a greater effect on performance. However, we did not observe any trend where larger $\gamma$ consistently performed better. Thus, additional study is required to validate or disprove our hypothesis. There are several directions in which this work could proceed. The first is to consider other domains, particularly settings with longer time dependencies. It may be that in the settings we studied, all of the relevant information could be captured by short timescale dependencies and longer ones added no additional information. Also, studying our hypothesis with the architecture in this paper may have been complicated by the use of the GRU memory unit that was employed. It may be clearer to study simpler architectures which lack explicit memory structures.

Instead, our primary contribution is the observation that adding auxiliary tasks allows us to shorten the trajectory length used in the A2C update. Shortening the trajectories may be useful for enabling the agent to update its policy more frequently, adapting quicker to changes in the environment.
We believe this represents a general trend that auxiliary tasks can make policy learning less sensitive to parameter settings. This sort of reduced sensitivity, or increased robustness, was previously reported by \citet{Jaderberg2017}; this work supports that observation.

One of the unanswered questions from our experiments is why does learning fail when the trajectory length is shortened? Generally, as the trajectory length was shortened, performance degraded. However, it did not degrade smoothly, but instead appeared to fall into failure modes where, after a short period of learning, the policy was unable to improve further. 

Performance of the policy could be sensitive to the weighting placed on the auxiliary tasks. Thus, it would be beneficial to use an algorithm that automatically tunes the weighting parameters, such as the meta-RL algorithm employed by \citet{Xu2018}.



We close by restating our observations: 1) adding TD-AE tasks can improve the robustness of the A2C algorithm to its trajectory length parameter setting, 2) temporally extended TD-AE can sometimes help more than just an autoencoder, 3) performance of the policy is sensitive to the weighting placed on its loss, and 4) even poorly tuned weightings enhanced agent performance.




\bibliographystyle{ACM-Reference-Format}  
\bibliography{references.bib}  

\end{document}